# Generative AI and linguistic diversity in academic writing and publishing: Perspectives from World Englishes

Kingsley Ugwuanyi, SOAS University of London, UK
Christian Mair, University of Freiburg, Germany
Sender Dovchin, Curtin University, Australia
Iker Erdocia, Dublin City University, Ireland
Maria Kuteeva, Stockholm University, Sweden
Esther Airemionkhale, Carnegie Mellon University, US

## Abstract

The rise of generative artificial intelligence (GenAI) in academic writing and publishing (AWP) raises questions about linguistic inclusivity and the legitimacy of diverse Englishes in global scholarly communication. This article responds to these questions through a structured scholarly dialogue involving five sociolinguists from World Englishes and adjacent fields. Organised around five guiding questions, the dialogue interrogates how GenAI tools influence writing practices, reinforce or disrupt dominant language norms, and raise ethical challenges. Contributors reflect on the potential of GenAI to democratise writing processes while also raising concerns about GenAI's tendency to marginalise minoritised varieties and flatten nuance in scholarly writing. Across the dialogue, themes of linguistic (in)justice, researcher agency, and institutional responsibility emerge, with contributors calling for equity-informed policies, critical AI literacy, and inclusive co-design in GenAI development. The article shows the value of dialogic reflection in understanding GenAI's role in AWP. It concludes that while GenAI may reinforce existing hierarchies, it can also serve as a site of resistance, depending on how it is designed, governed and used within scholarly communities committed to linguistic diversity.

## 1. Introduction

One key concern arising from the ongoing exponential rise in GenAI technologies is its implications for linguistic variation in academic writing and publishing (AWP). While GenAI technologies offer new opportunities in AWP, they also pose concerns regarding their potential to homogenise linguistic styles, privilege dominant English varieties, and marginalise scholarly voices from diverse linguistic backgrounds. Such concerns are especially pertinent within the field of World Englishes, which fundamentally recognises the legitimacy, diversity, and value of English varieties beyond dominant norms. This article interrogates these emerging concerns through a structured 'scholarly dialogue' with sociolinguists from a World Englishes perspective. The dialogue is guided by five questions addressing GenAI's broad influence in AWP, its potential biases towards dominant Englishes, institutional responsibilities, peer review practices, and ethical frameworks for GenAI use in AWP. A recent study involving interviews with applied-linguistics journal editors reports policy ambiguity, and a cautious stance that limits acceptable GenAI use largely to language polishing, with transparency deemed essential (Moorhouse et al., 2025). This article differs in focus: it brings together World Englishes scholars (some of whom are/were also editors) to foreground linguistic diversity and justice, extending the discussion beyond policy compliance to questions of whose Englishes are (de)legitimised in AI-mediated AWP.

To draw diverse insights and experiences from scholars from a wide range of backgrounds, five scholars were purposively invited to contribute to the discussion. Efforts were made to invite scholars with diverse levels of experience, regional backgrounds, and research interests (though all engage with World Englishes in some way). Importantly, most of the contributors have already investigated aspects of the role of GenAI in the academic writing processes.

Prof. Christian Mair (University of Freiburg, Germany) brings extensive expertise in World Englishes, with a notable focus on language variation and globalisation. Prof. Mair is popularly known for articulating the concept of the "World System of Englishes," conceived as a framework for understanding how English varieties interact across transnational and digital realms (Mair, 2013). Dr Iker Erdocia (Dublin City University, Ireland) contributes insights at the intersection of sociolinguistics and linguistic policies, particularly around linguistic diversity and AI technologies, drawing insights from his recent work on ideologies of dataism in the age of AI (Erdocia et al., 2025). The next contributor, Prof. Sender Dovchin (Curtin University, Australia), enriches the dialogue through her extensive research into linguistic diversity, translanguaging, and global Englishes in digital communication. In her recent work in which she described GenAI as "a double-edged sword" (p. 410) in academic writing, she proposed the concept of "algorithmic colonialingualism" (Dovchin, 2024, p. 412), which she discusses further in this conversation. Prof. Maria Kuteeva (Stockholm University, Sweden), whose work intersects academic English, multilingualism, and linguistic variation in publishing contexts, is well known for her work in publishing in multilingual academic contexts, as well as her work in linguistic diversity in the age of AI (Kuteeva & Andersson, 2024). Finally, Esther Airemionkhale (Carnegie Mellon University, US), whose work investigates the complex roles that postcolonial Englishes play in identity formation, provides an essential 'doctoral voice' in the dialogue, adding an early-career perspective, which is often marginalised in scholarly discourse of this nature.

While the dialogue represents a range of regions, academic stages, and scholarly orientations, it must be acknowledged that all contributors are currently based in institutions located in the Global North. Nonetheless, Kingsley Ugwuanyi's and Esther Airemionkhale's background as Nigerians, coupled with Sender Dovchin's engagement with the linguistic situation in Mongolia, brings the needed Global South perspectives to the discussion. Still, the absence of voices based in Global South academic institutions, especially those of non-Anglophone traditions, reflects broader structural inequities in global AWP, which future dialogues of this kind should aim to address through more inclusive representation.

As the convenor of this dialogue, I draw on my background in World Englishes and sociolinguistics (with particular interest in postcolonial Englishes) to frame the questions and synthesis presented here. Trained in and working across both African and Global North academic contexts, my own position would inevitably shape how I interpret issues of linguistic diversity, equity, and GenAI use in AWP.

From the above, it is clear that the contributors are well positioned to share their insights regarding the place of GenAI in promoting or discouraging linguistic diversity in AWP. In order to retain the *integrity* of the scholarly voices in this dialogue, their responses are presented verbatim in the article, with analytical commentary provided to highlight areas of convergence and divergence, contextualise insights within broader debates in the field, and articulate implications for future AWP practices. The following section is categorised into five subsections to reflect the five questions posed to the contributors.

## 2. Scholarly dialogue on GenAI in scholarly writing and publishing

### 2.1 The evolving influence of GenAI in academic and publishing

*Question: In your experience or opinion, how is Generative AI (e.g., ChatGPT, Gemini, Claude, CoPilot) influencing academic writing and publishing practices today?*

Maria and Sender open this topic by overviewing the general impact of the use of AI technologies in AWP. As Sender put it, AI technology is “a double-edged sword” because it comes with “ambivalent outcomes” (Iker).

> **Maria:** In my experience as a reviewer and journal editor, GenAI has impacted academic writing and publishing practices in several ways. On the one hand, large language models (LLMs) have facilitated access to knowledge and genres that have long been considered occluded. This certainly helps to demystify the peer reviewing process. For example, with a simple prompt for the free version of ChatGPT-4: “What are the most common peer review comments from applied linguistics journals?”, a novice researcher receives a whole range of concerns that are typically expressed by reviewers with regard to theoretical and conceptual issues, methodology, data analysis, research contribution, ethics, and so forth. At the same time, the advice generated by the bot reflects the conventions and language norms that are typical of the Anglophone scholarship and related rhetorical conventions, e.g., emphasising the need to cite more recent studies and reviewing language to ensure clarity. But we know that these norms are not universal. While recent calls to decolonise applied linguistics scholarship underscore an epistemic bias towards Western modes of thinking, LLMs are driving the ways in which knowledge is produced and communicated further toward standardization and convergence (Kuteeva & Andersson, 2024).
>
> **Sender:** AI has become such a big thing in the last few years. I initially tried not to pay attention to it, but then I couldn’t anymore. When I saw that many research papers have started discussing it, I decided to read up on what GenAI is and how it is influencing academic writing and publishing. From this initial review of literature, I learned that academic writers have begun to use GenAI for a wide range of reasons. For example, Khalifa and Albadawy’s (2024) review outlines six ways in which academics are using GenAI: a) facilitating idea generation and research design; b) improving content and structuring; c) supporting literature review and synthesis; d) enhancing data management and coding; e) supporting editing, review and publishing; and f) assisting in communication outreach and ethical compliance. Then, I started asking my colleagues how much they use AI and for what purposes. From our conversations, I found that while the non-tech-savvy older scholars either do not use GenAI at all or use it sparingly, junior scholars tend to use it more and were more positive about its usefulness, some of them describing GenAI tools as co-researchers. To sum up, AI is starting to influence academic writing, but there remains some concern about the ethical issues around its use.

Sender's point above about there being a general gap is also echoed by Christian below, who says that he has yet to use AI tools for academic writing, but his students do. However, some young scholars, such as Esther (below), rarely use GenAI for her writing tasks for fear of losing her writer's voice. Despite these, there is emerging evidence that young people use AI tools more than older folk, "with over-55s using AI half as much as over-35s" (Milmo, 2025, n.p.). While using AI tools for writing might be uncommon among older scholars, it is possible they are using them for other tasks, such as translation, as Christian mentions below.

**Christian:** I have not used AI tools for data analysis and for my own academic writing so far. Students whose work I supervise are generally more willing than I am to experiment with outsourcing some of the data analysis to AI. I've seen them apply LLMs for topic modelling and data annotation, more or less successfully. Methodological and ethical conundrums arise all the time. In practical terms, I have saved a lot of time through AI-based machine translation, but I restrict this to languages that I speak fluently or at least well enough to assess the accuracy of the output. Generating and machine-translating high-stakes documents, such as letters of reference, peer reviews or assessments for promotion, remains highly problematical and potentially unethical in my view.

**Esther:** In academic writing and publishing, many scholars (regardless of their first language) now turn to GenAI to *standardize*, structure, and critically self-assess to increase the chances of journal acceptance. Some reviewers use GenAI to evaluate a manuscript's content, originality, plagiarism, and ethical use of AI. As a doctoral researcher, I am cautious about using GenAI in academic writing for several reasons. First, I think there is a reduced opportunity to critically engage and produce knowledge. Second, I fear that producing knowledge with AI strains my ability to find my writer's voice or silence it. More than ever in academic writing, I am more careful with word choices so as not to *sound* like GenAI. For example, if I use "delve" in my introduction or "underscore" in my concluding paragraph, I worry that my work may read like AI-generated content, so I reword it with the hope that the reviewer will see and acknowledge that it's a human, not an AI, language style. Why these words? In an article by Grammarly (2025), it was noted that these two words, which were arguably once considered very scholarly, are, in fact, commonly overused by AI. Similarly, in a tweet, Nguyen (2024) gave a statistical overview of the drastic increase in the use of "delve" in *PubMed*, a medical journal. Beyond academia, Graham (2024), a tech entrepreneur, argued in a tweet that using "delve" in an email (or scholarly article) increases the likelihood that the content was generated by AI. So, what does that mean for doctoral researchers who really want to *delve* into a topic? Do reviewers have assumptions about these high-frequency AI words when they come across them in manuscripts? How might these assumptions impact doctoral researchers' chances of publication? For early career researchers, one certainly does not want to have an AI style, but what does it mean to sound like AI? We often forget that humans wrote the dataset that LLMs are fed with, and somehow, we judge GenAI as a better writer.

Esther's anxiety about "sounding like AI" might not simply be about writing competence but about maintaining epistemic sovereignty in academic voice, similar to what Bisht's (2023) concept of decolonial literacy. A further consideration of this view from a decolonial lens supports Canagarajah's (2024) argument that multilingual scholars negotiate dominant publishing norms while asserting alternative rhetorical identities in AWP. Iker then broadens the conversation by approaching GenAI in AWP from the perspectives of access and equity.

**Iker:** GenAI is a technology with ambivalent outcomes. It opens a range of affordances, enabling researchers to analyse data, conduct literature reviews and perform more complex tasks. From a linguistic perspective, it is good at paraphrasing, suggesting discipline-specific vocabulary, editing for standard grammar, translating, etc. Considering the writing conventions and expectations of academic registers and genres, conforming to an idealised standard in writing can be challenging for many speakers of non-standard varieties of English. In this regard, AI tools can be particularly helpful for early-career and multilingual researchers, and those with limited access to institutional resources and support systems for publishing. Some people tend to present this as a democratisation of data access and knowledge generation, claiming that it opens the way to publishing opportunities. Part of it may be true, as there are other non-AI-related gatekeeping mechanisms in academic publishing (Pérez-Milans, Kaufhold & Soler, 2025). I remain agnostic regarding these democratisation claims. This is because GenAI raises serious concerns about authorship, originality and academic integrity, not to mention issues around copyright, monopolistic practices, exploitation and environmental costs.

The clearest lines of thought running through these initial viewpoints are that: a) AI technology can no longer be ignored in AWP, b) it comes with both positives and negatives, and c) it raises ethical concerns relating to authenticity and equity. Interestingly, all commentators approached these issues from different perspectives, which helps bring balance to these opening discussions about the emerging role of GenAI in AWP. These reflections underline the need for a decolonial approach that treats engagement with AI not as mere adaptation to new technologies, but as a critical negotiation of voice, authority, and epistemic justice in global academic publishing.

## 2.2 GenAI's handling of World Englishes in scholarly writing

*Question: Given that most GenAI tools are trained on dominant Englishes, how do you see this shaping the way these tools treat World Englishes in academic contexts? What concerns or possibilities does this raise for the visibility and legitimacy of diverse English varieties in research publication?*

A recurring tension in the expansion of GenAI is its simultaneous potential to democratise access to academic publishing (as echoed in Iker's response above) and reinforce global linguistic hierarchies. The responses in this section focus on this paradox. While contributors recognise that GenAI can facilitate academic writing, they raise serious concerns about how its standardising tendencies threaten the diversity of norms enshrined in World Englishes scholarship. This is what Sender calls GenAI's "potential to flatten nuances" through outright erasure, misrepresentation, or caricature. Maria opens this topic with an example from her experience of prompting ChatGPT-4 to produce a response in a non-dominant variety of English, and the outcome is outrageous.

> **Maria:** At this point, even the most recent versions of LLMs function as a statistical predictor, labelled 'a stochastic parrot' by computational linguists (Bender et al., 2021). Below is a diagram generated by ChaptGPT-4 to illustrate the process of answering my query "What are the most common peer review comments from applied linguistics journals?", mentioned in Question 1:

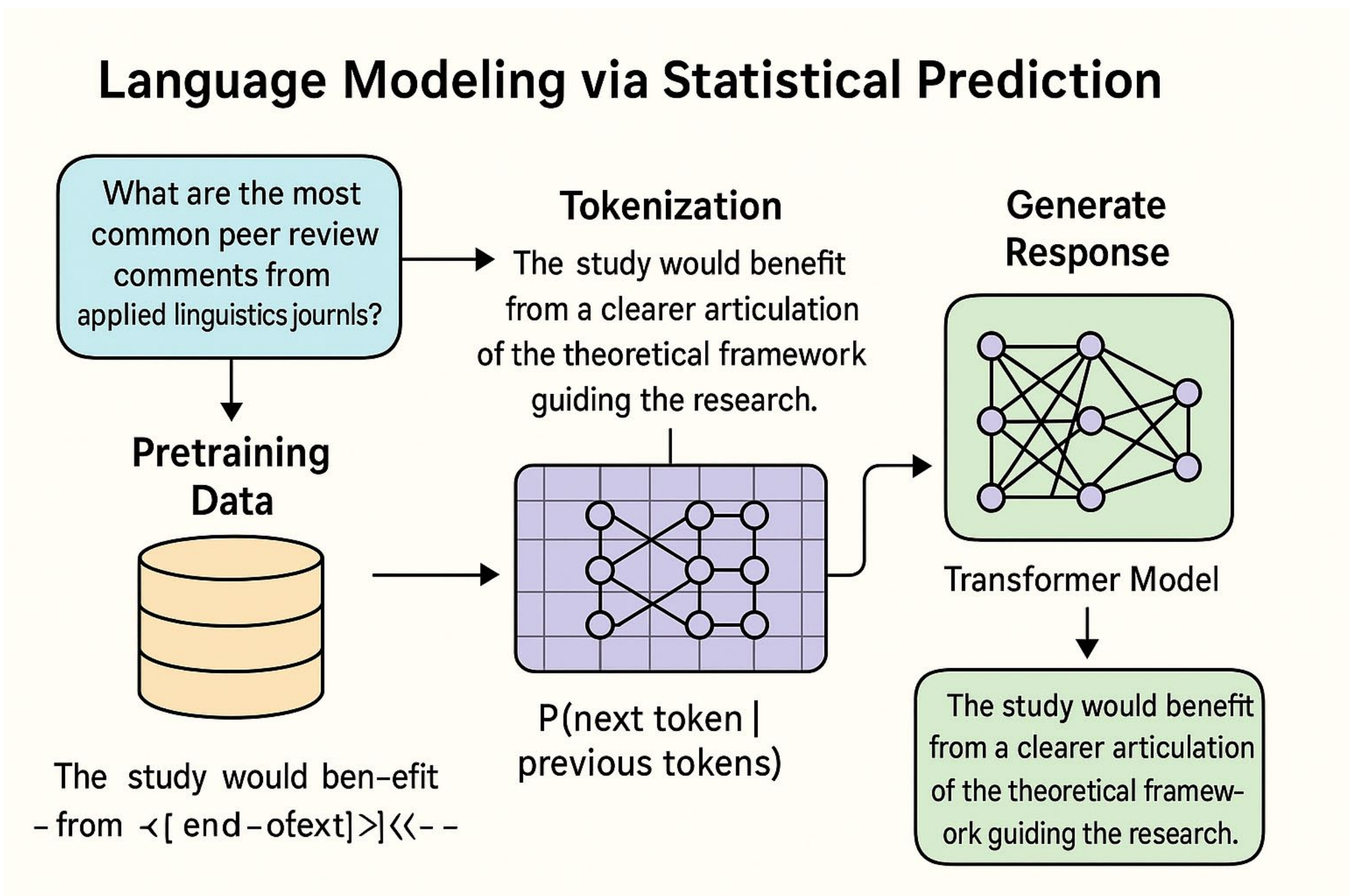

Fig. 1. ChatGPT-4 output illustrating the process of answering the query about the most common peer review comments from applied linguistics journals.

Fig. 1 shows how an LLM statistically predicts text. Although the example answer sounds like a real comment, the model is simply calculating which words are most likely to follow one another based on patterns in the texts included in its corpus. In order to answer the question in the prompt, the model does not look up any actual peer reviewer reports. Rather, it recognizes patterns from the data used during its training, e.g., scholarly discussions, reviews, etc. With regard to pretraining data, an LLM learnt statistical relationships between words and phrases, broken into tokens, to recall most likely probabilities (e.g., after "the study would", the word "benefit" is most common). The transformer model predicts the next most statistically likely token, building a full sentence: "The study would benefit from a clearer articulation of the theoretical framework guiding the research." This output indeed sounds like a common peer reviewer comment, but is difficult to interpret when applied to any specific manuscript as it lacks contextualisation.

When it comes to generating texts in English, inbuilt biases start already at a pre-training stage which affects content (Bender et al., 2021; Payne et al., 2024), as the data used tends to be collected from Internet sources that overrepresent younger (male) populations based in the so-called WEIRD (Western, educated, industrialised, rich, and democratic) countries (e.g. Reddit, Wikipedia, etc.). Thus, LLMs are predominantly trained on specific registers of English and cultural references (Payne et al., 2024). They struggle to deal with other ways of thinking or generate text in registers or languages that it has not been exposed to.

LLMs also tend to reproduce standard or dominant varieties of language and marginalise other varieties, including the World Englishes used outside the Kachruvian inner circle. When an LLM reproduces a non-dominant variety of English, it tends to exaggerate its linguistic features, reinforcing stereotypes (Jones, 2025). Below is an excerpt from ChatGPT-4's response to a one-shot prompt to rewrite the abstract for Kuteeva & Andersson (2024) in Nigerian English:

**Nigerian English version (by ChatGPT):**

People wey dey do research for different areas don dey talk say e good make we diversify and remove colonial mindset from the way we dey share and produce knowledge. True-true, these big-big AI language tools like LLMs fit help people access wetin dem don write for abroad, and even show how dem dey publish academic work. But still, these LLMs get their own wahala. Dem still dey follow the same old pattern wey favour the kind knowledge wey come from abroad, and dem no sabi local matter or how human beings really dey use language.

Leaving aside the limitations of LLM output generated in response to one-shot prompts, the above text generated by ChatGPT confirms that the voices of scholars using minoritised varieties of World Englishes are clearly underrepresented in the pre-training data. The linguistic features used in the generated text exaggerate the speech styles associated with Nigerian English that have been enregistered in digital media to such an extent that the generated text would no longer be considered academic. To this day, LLMs tend to essentialise local cultures (Jones, 2025) and are not flexible enough to produce a nuanced output that reflects local features without falling into a stereotype.

> **Christian**: From a World Englishes point of view, this question goes to the very heart of the matter. To put it bluntly: If you look at the formal/standard end of the style continuum in LLM output, the plural *dominant Englishes* in your question is not appropriate, as only *one dominant English* remains, namely the US standard variety. I have carried out prompting experiments to generate text in national standard varieties of English other than the US one and found LLMs to perform very badly at this task. This even affects British English. AI will offer to produce output in British English when prompted, but in my experience it cannot even implement British spelling consistently over more than a few paragraphs, let alone produce the full lexicogrammatical profile of British English text. Whether it is our current notion of "national standard varieties of English" that needs revision or whether the training data supporting LLMs are biassed against them is a question that I would not want to answer definitively yet. In a situation in which the future is bleak for (written) British Standard English, still a globally relevant reference variety in the offline world, what are the chances of less dominant, less well-documented and less strongly codified standards in the pluricentric global English Language Complex? It seems that LLMs confront us with a standardisation paradox, with English becoming more homogeneous and more heterogeneous at the same time. This paradox has the interesting result that, if you prompt for a Nigerian English or a Jamaican English version of a text written in American English, you will definitely not end up with output that is *Nigerian/Jamaican Standard English* as described in handbooks, but stuff that has lots of *Nigerian Pidgin* and *Jamaican Creole* in it.

Both Maria and Christian show how GenAI is biased against non-dominant Englishes. As Christian points out, it is bewildering that even British English seems to be marginalised by AI technology, as the monopoly of standard American English reigns supreme in GenAI, which is supported by the finding that responses generated by ChatGPT almost immediately revert to American English, sometimes by a margin of over 60% (Fleisig et al., 2024). Both responses, as well as those below, trace the source of these biases to the training data. One interesting finding from the examples of Maria and Christian is that ChatGPT does not differentiate between Standard Nigerian English and Nigerian Pidgin (which is a different language altogether), which, according to Christian, raises several foundational questions about how we define national varieties of English. Christian complicates the issue further by introducing what he calls a "standardisation paradox," where GenAI both homogenises English but also opens new opportunities for hybridised forms, a point that anticipates Esther's points below.

> **Esther**: Technology is not neutral, and different GenAI tools serve different purposes. With that in mind, the affordances and constraints of specific GenAI tools for World Englishes depend on how they are used for writing. I often use Grammarly, a spelling and grammar checker, for academic writing. The premium version I use is integrated with GenAI; sometimes, there's a tug-of-war over the appropriateness of a word, phrase, or expression. The good thing is that one can always ignore all suggestions. Currently, Grammarly has one non-inner circle English (Indian English) as one of the options for English. While I do not know the extent to which Indian English can be expressly used within Grammarly, its presence on the list of varieties shows AI writing assistants' (Grammarly, Wordtune, etc.) potential to legitimize World Englishes. For instance, while doing this reflection, I decided to test two GenAIs on variation: Assistant, an LLM within Poe, and ChatGPT. I copied a paragraph from my existing work and used the following prompt: *Please help me rewrite this in Nigerian English [insert paragraph].* Unsurprisingly, Assistant rewrote the paragraph in Nigerian Pidgin (NigP) as if that were Nigerian English. While misconstruing Nigerian English as NigP is not unusual among humans in the real world, its existence in GenAI is a step towards visibility.

**Iker:** As currently developed, LLMs reproduce linguistic hierarchies: they are built on dominant standard English varieties, often mainstream American English, and produce a monolithic version of English that ignores variation across other regional varieties of English. For example, ChatGPT's writing suggestions tend to convert Indian English into American English (Agarwal et al., 2025). In a more positive tone, LLMs can be fine-tuned to improve the performance across languages and in varieties of the same language, including non-standard forms (Faisal & Anastasopoulos, 2024). As with efforts to address gender bias in AI, optimised models and multilingual language models could help mitigate the systemic linguistic bias towards Anglophone centres and accommodate linguistic variation (cf. Hofmann et al., 2024). But to make progress, we need concerted interventions to develop AI governance frameworks that take linguistic diversity seriously. Further, we should not overlook the agentive capacity of individual researchers to edit or completely rewrite GenAI-generated outputs before submission, for example, in ways that reflect their linguistic identity. That said, AI-enabled tools could reinforce gatekeeping practices inherent in the academic publishing system by, for example, misinterpreting grammatically systematic and rule-governed constructions within a given English variety and perpetuating epistemological paradigms connected to academic centres in the Western world and, more generally, the coloniality of knowledge production. We need empirical research, including ethnographic-oriented approaches, to better assess the impact of GenAI on academic publishing (e.g., Qijia & Shikui, 2025).

Where Esther reflects on user agency in adapting GenAI outputs, Iker foregrounds the interventions required to make such agency meaningful. He acknowledges the potential for fine-tuning and multilingual optimisation but warns that these remain unrealised without deliberate governance frameworks. Sender closes the dialogue with an insightful reframing of linguistic marginalisation in GenAI as "algorithmic colonialingualism," which she says leads to the exclusion of the linguistic repertoires of marginalised speakers (such as Indigenous peoples) from AI design and functionality. Drawing from translanguaging theory, she illustrates how GenAI fails to accommodate flexible, hybrid practices that characterise everyday communication for many multilingual speakers. Her call for co-designing AI systems with marginalised communities anchors the conversation in inclusion and justice, bringing the dialogue full circle to the questions of legitimacy and epistemic control raised by other contributors.

> **Sender**: I had the opportunity to write an afterword for a special issue in the *Australian Review of Applied Linguistics* (Dovchin, 2024). After reading all the special issue articles, my conclusion was that artificial intelligence is a double-edged sword. On the one hand, it's really revolutionising everything. On the other hand, I was concerned that AI has the potential to colonise the language again. In that paper, I said that GAI might give rise to what I call "algorithmic colonialingualism." What I mean by this is that the algorithmic system of GenAI is based on standard English varieties, while most varieties of world Englishes are not often reflected. For example, Aboriginal English spoken by First Nations people in Australia is not reflected in artificial intelligence. Further, my research centres on translanguaging practices. We know that when translanguagers speak on a daily basis, they move fluidly and flexibly between different linguistic and semiotic repertoires. For example, if you try to analyse translanguaging data with the help of AI, it's not just going to work, because the AI system is not designed for that. To ensure that GenAI systems are more inclusive, their algorithmic systems should be co-designed with people from Indigenous backgrounds, migrant/refugee backgrounds, and people from the Global South.

With these responses raising questions about the inbuilt biases in AI (Maria and Christian), user agency (Esther and Iker), and colonialingualism (Sender), a recurring theme across all five responses is that GenAI privileges dominant norms (particularly Standard American English at the expense of other Englishes. This dominance exemplifies what Tupas (2015) terms "unequal Englishes," where certain Englishes gain epistemic legitimacy while others are reduced to caricature or error. From a World Englishes standpoint, GenAI's apparent universality therefore masks a new phase of linguistic imperialism. Yet there is also cautious hope: the possibility of intervention through user resistance, technological fine-tuning or inclusive co-design suggests that GenAI's linguistic future is open to many possibilities.

### 2.3 Institutional roles in safeguarding linguistic diversity in the age of GenAI

*Question: Do you think academic institutions, journals, and publishers have a responsibility in ensuring that the use of GenAI supports, rather than undermines, linguistic diversity? If yes, what might this responsibility be, and how do you think it could be put into practice? Relatedly, do you think journals/publishers should provide guidance for peer reviewers who may be using GenAI tools to evaluate the language or content of submissions, especially those written in non-dominant Englishes?*

As the use of GenAI tools becomes increasingly embedded in academic research and publishing, questions about institutional responsibility have been raised. In this section, contributors reflect on what academic institutions, journals, and publishers can and should do to ensure that GenAI supports rather than undermines linguistic diversity and scholarly inclusivity.

**Iker**: Academic institutions and publishers have a key responsibility in ensuring fair accessibility to the publishing system, which is currently structured around the supremacy of English. Constructive discussions in editorship in sociolinguistics and related fields are taking place (e.g., Piller, 2022), yet inconsistencies often emerge between the stated ethos and actual practices (Soler, Erdocia & Savski, 2023). For example, some journal guidelines impose an ambiguous standard of 'good English', sometimes aligned with either American or British English (McKinley & Rose, 2018). These guidelines often position L2 and other writers of non-dominant Englishes as deficient of legitimate standards, which raises ethical considerations of accessibility and inclusivity in publishing. The traditional academic system failed to evolve in ways that properly addressed linguistic diversity long before the AI era. Since current AI models are biased toward dominant norms, it is unlikely that they will contribute to reversing the lack of variation and other systemic inequalities in mainstream knowledge-making practices (Kuteeva & Andersson, 2024). Options to raise awareness and ensure inclusive practices among reviewers, editors and PhD supervisors and examiners include developing inclusive guidelines for linguistic variation, promoting critical GenAI literacy and providing means for editing purposes where necessary. In my role as faculty delegate for research integrity in my institution, I see firsthand how universities and publishers are striving to keep pace with AI-related developments from a regulatory perspective. This could be a timely opportunity for us to advocate for real change in linguistic inclusivity and diversity.

**Esther**: Academic institutions and publishing bodies should be aware of the limitations and biases in LLM language norms. Reviewers using GenAI in their reviewing tasks should be critically aware of variations in World Englishes - just because it is written differently does not mean it is wrong. Hence, stakeholders can provide resources for reviewers to guide and familiarize themselves with World Englishes. The current homogenous English language norms in academia position GenAI as a leveler in academic writing. The challenge is that if scholars are not writing in what is considered authentic to them, how can we present non-Eurocentric forms of knowledge without succumbing to Eurocentric ways of seeing the world?

**Christian**: Generally, the answer to this question will have to be negotiated between individual research communities and their preferred publication outlets. In my view, a journal on World Englishes not recognising the plurality of standards would be outrageous, and I would expect a generous and inclusive policy on this point among any journal dealing with linguistics, literature and cultural studies. As for ESL speakers, I wonder whether their nondominant varieties of English are merely the tip of an iceberg and that the really harsh discrimination against them runs deeper. Scholars in many ESL countries suffer from lack of access to up-to-date research literature. This prevents them presenting their arguments at state-of-the-art level. Also, research infrastructure, for example linguistic corpora, and research agendas are often created in the Global North.

Iker, Esther and Christian all show that the academic institutions and gatekeepers have a role to play in upholding diversity in Englishes in AWP, particularly in this era of AI. As Iker points out, the exclusion of non-dominant Englishes in AWP predates the arrival of AI, as what is traditionally regarded as "academic English" is based on the norms of standard varieties of the inner circle (Phillipson, 1992). Christian also argues that the exclusion of the norms of non-dominant Englishes is only one aspect of the inequities experienced by scholars from the Global South, calling for institutions to pay attention to addressing these. Sender and Maria advance the discussion by proposing equity-informed GenAI policies across academia. Both of them also emphasise the need for transparency in the use of AI in AWP, arguing that while academic institutions should provide policies that encourage linguistic inclusivity, the onus also lies on academics who use AI technologies to declare when, and the extent to which, they use them. In fact, there is what is called an AID [Artificial Intelligence Disclosure] Statement in AWP in which researchers are encouraged to declare their AI use (Weaver, 2024).

**Sender**: The majority of academic institutions, journals, and publishers are still in the process of sorting out how to approach the use of AI. From my own experience, until I completed my tenure as the editor of the *Australian Review of Applied Linguistics* about two years ago, there was no policy on AI. I think publishers and academic institutions are as concerned as we are, so we should all work together (including with relevant government agencies) to co-design these policies. What I think is that these institutions should be really supportive of using non-standard world English varieties and translanguaging practices in artificial intelligence. Scholars from indigenous, migrant, and global south backgrounds who are using artificial intelligence should be able to see that their way of speaking is reflected in AI systems. I think institutions should be very clear that they have equity-informed artificial intelligence policies that will encourage writers and reviewers to just avoid the automatic standardisation of English, for example, and consider the context and divergent voices and linguistic identities when using AI. Further, I think journals and publishers should ask writers to declare the extent, if at all, of the use of AI in writing. Some journals are already doing this. I have also noticed that some journals now tell reviewers not to use AI in peer-reviewing. This is because AI-assisted assessment has the potential to flatten nuances and misinterpret culturally grounded expressions. For example, what if the paper is about translanguaging, and then the peer reviewer is going to put the paper into AI? It's absolutely not going to make any sense. Reviewing academic papers using AI will be harmful to academic writing, especially to scholars from the Global South, those with non-English speaking backgrounds, and those working with marginalised populations.

**Maria**: Everyone involved in the publishing enterprise is responsible for ensuring that GenAI is used ethically and in line with best research practices, and that its use is disclosed. When it comes to language use, as mentioned above, LLMs are unlikely to

> contribute to supporting linguistic diversity, at least not at this stage. Although they are widely used in the academy, the decisions pertaining to what to include in pretraining data and what filters to use when selecting/generating content are made by large corporations in the US and China. Kuteeva & Andersson (2024) provide a few suggestions on how to resist the trend toward knowledge and language standardization.

The dialogue in this section reveals a collective acknowledgement that institutions, journals, and publishers cannot remain neutral bystanders in the GenAI era. All five contributors agree that action is required. Contributors propose institutional guidelines that affirm linguistic diversity, policies that ensure reviewers understand variation within English, and declarations of AI use in writing and peer review. Editors themselves describe unclear GenAI policies and call for discipline-specific guidance, in order to ensure that institutional responsibility is clear (Moorhouse et al., 2025). These findings strengthen the case for reviewer education on linguistic variation alongside explicit GenAI policies for institutional stakeholders. While no single measure will suffice, the consensus is clear: ensuring GenAI supports inclusivity will require intentional, sustained, and collective action from all stakeholders involved in AWP. Taken together, the stance taken by the contributors here resonates with Saulière's (2014) notion of sociolinguistic responsibility, which situates linguistic justice within the ethical obligations of institutions rather than just individuals. In this way, institutional engagement with GenAI becomes less about compliance and more about cultivating ethical spaces where plural Englishes can coexist.

### 2.4 Quality of academic writing since GenAI

*Question: From your experience as a peer reviewer or journal/book editor (if applicable), have you noticed any changes in the overall quality of manuscripts since the advent of GenAI? If so, how would you describe those changes, and what do you think they might mean for the future of scholarly writing and linguistic diversity in publishing?*

As GenAI becomes more integrated into academic workflows, its effect on the quality of scholarly writing is increasingly scrutinised. While most contributors say they have noticed some changes in the quality of scholarly writing since AI, a few of them think AI's impact on the quality of writing has not been phenomenal.

> **Christian**: The first part of the question is easy. Yes, the quality has improved, as many authors seem to have their drafts proofread and optimised in other ways. If this is done in the interest of editorial consistency and clarity, that is fine with me. If I sense "padding" in the form of nice-sounding but ultimately trivial generalities, I find it irritating.
>
> **Maria**: There are now fewer manuscripts that would be sent back to authors for proof-reading and language checking. Some manuscripts acknowledge the use of AI, e.g. for statistical analyses. On the whole, it seems that many authors use GenAI one way or another but do not necessarily declare for whatever purpose. From the perspective of journal editors, one tangible result of this trend is an increased number of submissions. The sheer quantity of manuscripts written in what appears to be correct English does not lead to better quality of research. Instead, it further complicates an already challenging context in which editors need to work even harder to screen submissions for eventual desk rejection and to find expert reviewers for the remaining submissions.

Although the increasing linguistic correctness observed by Christian and Maria might seem to signal an improvement in academic writing, Maria explicitly cautions that this "sheer quantity of manuscripts written in what appears to be correct English does not lead to better quality of research." In that sense, her observation aligns with Iker's and Esther's later reflections that apparent improvement often masks deeper homogenisation, which undermines linguistic diversity and flattens cultural nuance (Dai & Hua, 2025; Agarwal et al., 2025). Large-scale corpus evidence now shows a marked rise in AI-associated style words in linguistics abstracts (e.g., delve, pivotal, intricate), suggesting increased LLM mediation in linguistics research, and a drift towards uniform stylistic registers, which raises transparency and equity concerns even when texts appear "improved" (Botes et al., 2025). While Botes et al. treat word-frequency shifts as indirect indicators rather than proof of AI authorship, the pattern remains strong enough to warrant concern about creeping homogenisation, which challenges the very essence of the world Englishes paradigm.

**Iker**: As an author, I have come across reviewers' comments noting some oddities in my manuscripts. Some are unintended influences from Spanish, my first language, on my English writing. Others reflect my personal style and are a conscious decision about what I consider the best way to convey my points and express myself academically. But some reviewers, invested with the power of anonymity and native-speakerism, do not seem to concur with some of my choices. One anecdotal example is my use of the Irish English phrasal verb "avail of something" without a reflexive pronoun. Both AI writing assistants and human reviewers diligently flag this use, suggesting instead the grammatically acceptable form in dominant English variants "they avail themselves of", which is strange to me as I am based in Ireland! At a granular level of text production, many term papers, manuscripts and abstracts, especially those written by early career researchers, seem to have been supported by Gen-AI, particularly when they overuse typical ChatGPT-style vocabulary and recurring syntactic patterns like dash-separated clauses and –ing elaborating clauses. While these texts are register-compliant and norm-conforming, they tend to be somewhat formulaic and lack flair. In our performance-focused culture, however, those texts often get the job done.

**Esther**: As I noted earlier, we are reinventing language as we write to sustain diversity in language and style. Yet, to what extent will (or not) such diversity be valued if GenAI's perfect language (and feedback) becomes the mark? I wonder if GenAI will dictate our writing benchmarks, with scholars striving to meet the mark it sets, or if scholars will be comfortable and confident in writing in their varieties. The originality and authenticity of our work face greater scrutiny because we have to worry about writing in a language appropriate for the academic genre and hope that humans and AI perceive us to have the capacity to produce the quality of

> language used. For instance, I recently ran a paper I wrote without the help of AI through an AI text detector before submitting it to a journal. Because I assumed the reviewers might do the same, I wanted to see what they might see if they looked. The result was ridiculous: AI returned an unbelievably high AI-detection score for a text I know I wrote myself. While there was nothing much I could do, this makes me question the fairness of evaluating quality and authenticity with GenAI.

Finally, Sender brings a candid note of hesitation, as she has not noticed a marked change in writing quality. She attributes this to the possibility that academics use GenAI more for brainstorming and structuring and perhaps less for composing full texts, as often assumed. She also thinks that it could be a discipline-specific issue, as applied linguistics, especially translanguaging data, which she works with, does not readily yield itself to AI use, given the contextual nuances that AI struggle to handle.

> **Sender**: To be honest, I wouldn't say that I've noticed that the quality of papers I review for journals has become much better now than a decade ago. It's still a similar voice and the fundamental academic writing style. So it's either there hasn't been a significant change in the quality of academic writing AI or, perhaps I've not been very observant. Maybe I should pay attention next time because I haven't really paid closer attention to it. It's also possible that academics are not using AI as much as we think. Or perhaps academics use AI more for brainstorming and structuring than they use it for actual writing. Another factor might be that in applied linguistics or sociolinguistics research, it is very hard to use AI for writing because we deal with very tricky data, such as conversations with far too many contextual nuances which AI cannot handle.

This section reveals that while GenAI is shaping the surface features of academic writing, its deeper influence on authorial voice, diversity, and scholarly values remains uncertain. Blommaert's (2010) notion of "orders of indexicality" reminds us that language varieties are not evaluated neutrally but through hierarchies of value that privilege certain forms as more legitimate. GenAI reinforces these hierarchies by automating what counts as 'good' writing. In other words, what appears as improvement in writing quality may therefore conceal the algorithmic enforcement of standardised norms. These tensions raise urgent questions: Are we heading towards a scholarly publishing landscape where variation is penalised, creativity and nuance are flattened, and quality is measured by algorithmic resemblance?

### 2.5 Principles for ethical and inclusive use of GenAI in academic writing and publishing

*Question: Finally, what principles do you think should guide the ethical and inclusive use of GenAI in academic writing—particularly in affirming the legitimacy of diverse Englishes in global scholarly writing and publishing?*

In this final section, contributors outline core principles that should guide how GenAI is used. While they differ in emphasis, all contributors converge on the importance of reflexivity, human agency, equity and further research, calling for a critical recalibration of our relationship with GenAI in AWP.

> **Iker**: In our advocacy role for linguistic diversity in academic writing, we must engage with the general language and discourses of ethical and responsible use of AI in research and, more specifically, with the principles of fairness and respect for users' agency. For example, it is difficult not to agree with statements such as "to achieve Trustworthy AI, we must enable inclusion and diversity throughout the entire AI system's life cycle … this also entails ensuring equal access through inclusive design processes as well as equal treatment" (European Commission, 2019, p. 18). However, we need to push for a linguistically oriented understanding of inclusivity and diversity in research policy making. Additionally, we should elaborate on what such a linguistic understanding means for related values such as transparency, agency consent and participatory and auditable governance in AI. To close, some variants of English are stigmatised as having a lower status and are, therefore, considered less legitimate for academic purposes. But we know well that language-related matters are not just about language. 'World Englishes' implies an unequal relation among speakers (Tupas, 2015), which is often entwined with colonisation and globalisation. These two are critical processes for an in-depth understanding of language and the ideologies of dataism in the age of AI (Erdocia, Migge & Schneider, 2025). The ethical dimensions and implications of AI have received considerable scholarly attention, with comparatively little emphasis on questions of power relations and institutional hierarchies. This calls for a more comprehensive approach.

> **Esther**: In academia, I believe these three core principles must guide our use of GenAI as writers or reviewers: reflexivity, accountability, and language awareness. Fundamentally, the biases and limitations of GenAI must be recognized to advance ethical and inclusive practices. We know that GenAI does not exist in a vacuum, and GenAI cannot be held accountable for how it is being used. With that established, we must continually and reflexively examine how we train, prompt or engage with GenAI in writing. Scholars must (re)evaluate how they define and practice fairness using GenAI while remaining committed to inclusive practices in the face of normative language ideologies.

> **Maria**: Based on my discussion above, the legitimacy of diverse Englishes in academic writing and publishing is, above all, a human endeavour. It is up to individual scholars to make sure that their voices and stances are not erased by an LLM in the process of language correction. This is where correct and nuanced prompts used in an iterative process of human-GenAI interaction can make a difference. It is important to raise critical awareness of the consequences of overrelying on LLMs in an attempt to write in standard varieties of English at the expense of promoting or at least maintaining epistemic diversity.

**Sender**: In that afterword that I wrote for that special issue (Sender, 2024), I said that we need to work towards a 'linguistically just' AI future, and that is the ethical principle for me. What I mean by this is that we need to have a collective commitment by policymakers, researchers and technologists to ethically co-design and co-implement these algorithms to ensure that AI systems fully support linguistic and cultural diversity (including from non-Western, Indigenous and global south knowledges) so everyone can benefit from it. But as I said earlier, AI is a double-edged sword. We always have to be cautious and critical about how we are going to approach AI because we don't really want to be colonised by robots and computers, because we really need to prioritise the human voice.

Christian closes the dialogue with a wide-angled ethical framework rooted in core scholarly values. He offers three principles: recognising that AI and ELF have come to stay in academia, resisting the temptation of superficial productivity, and reaffirming the centrality of human judgement, which is also echoed by other contributors, as shown above. One interesting dimension that he emphasises is the environmental cost of GenAI systems (de Fine Licht, 2025), which is an aspect of ethical considerations often neglected in academic discussions of the ethics of AI.

**Christian**: I would argue that there are three truths about AI that everybody working in the humanities and social sciences should hold to be self-evident:

(1) AI/LLMs and English as a Lingua Franca (ELF) are here to stay, as two "inevitable presences" in 21st century academia:

The new language technologies can be "domesticated" to some extent, but will be increasingly difficult to regulate or control through top-down policies. I expect a proliferation of increasingly detailed and complex guidelines, codes of conduct, etc., many of which will be out of date by the time they are approved. What we need instead is a simple and robust compass, in the form of a consensus about core values in research ethics and the resulting responsible scholarly use of the new technologies.

(2) All stakeholders, the weak and the strong, face the temptation of the path of least resistance:

For doctoral students and professors, for first-time authors and global players in academic publishing, for humble application writers as well as review boards dispensing millions of funding, AI presents a temptation to put quantity before quality. If authors, reviewers and publishers collude in this dystopian enterprise, the academic world will be swamped with unoriginal and superficial publications.

(3) We as scholars in the humanities and social sciences should follow AI developments in a spirit of constructive criticism, sharing in the celebration of what can be achieved, but continuing to take a stand for humanity.

> Human beings learn their native languages on the basis of a few hundred thousand words of input, perfectly, in a very short time, and with negligible negative impact on the environment. LLMs simulate these skills impressively, at considerable human cost and with enormous carbon footprint (Crawford, 2021). In this situation, we have every right to ask questions about developers' and corporations' priorities and ultimate agenda (which – in some extremer scenarios – might well be replacing human teaching and research by AI agents), but we also have a duty to use the new opportunities responsibly ourselves. Current debates on Responsible AI still tend to focus on two stakeholders, corporations and developers; however, we as academic users are equally called on to accept our responsibility. Discussing the important applied-linguistic and sociolinguistic dimensions of AI among ourselves is a necessary first step, but clearly not enough. We need to articulate our concerns in research cooperation with language engineers and developers, starting in our own institutions. And we need to integrate these new topics systematically into our teaching. Linguistics students with advanced digital and data literacy will have additional opportunities on the job market and help disseminate linguistic expertise in areas of language technology where it is needed.

In sum, these viewpoints emphasise the need to humanise GenAI use and to reframe AI not as a neutral tool for enhancing AWP but as a system embedded in histories, ideologies, and global asymmetries. They also argue that ethical considerations when using AI in AWP are not just about avoiding harm and declaring AI use in academic writing; it is also about intentionally creating space for diverse Englishes, epistemologies, and communities within academic publishing. This stance calls attention to the call for epistemic justice (i.e., the moral imperative to recognise and include marginalised voices in processes of knowledge production) (Fricker, 2007). Likewise, Crawford (2021) reminds us that AI ethics must go beyond technical design to ask questions of power, responsibility, and human consequence. In a world where AI is increasingly positioned as the arbiter of linguistic correctness, the collective insistence here is clear: humans must take the lead in the ethical campaign with humility, awareness, and commitment to justice and inclusivity.

### 3. Discussion and conclusion

This scholarly dialogue examined whether generative artificial intelligence (GenAI) tools support or undermine linguistic diversity and scholarly inclusivity, particularly from a World Englishes standpoint. Guided by five questions posed to five scholars, this dialogue interrogates many of the hopes, tensions, and unresolved questions raised in the scholarly discussions of the emerging role of AI technologies in AWP. Across all five questions, a central theme emerges: GenAI reflects and reinforces entrenched linguistic hierarchies, but it may also offer openings for resistance, reflexivity, and reform. As Iker put it, current LLMs "reproduce linguistic hierarchies…[because] they are built on dominant standard English, often mainstream American English," with the result that they "produce a monolithic version of English that ignores variation across other regional varieties of English." Similarly, Maria contends that the "voices of scholars using minoritised varieties of World Englishes are clearly underrepresented in the pretraining data," often leading to exaggerated or stereotyped outputs that undermine academic legitimacy (Fleisig et al., 2024).

The dialogue also foregrounds the standardisation paradox at the heart of GenAI in AWP. Christian captures this succinctly: "LLMs confront us with a standardisation paradox, with English becoming more homogeneous and more heterogeneous at the same time." The contributors acknowledge this dual role of AI. On the one hand, there are signs of the potential of AI technologies to accommodate several linguistic repertoires (Tafazoli, 2024), but they point out that a lot more work still needs to be done before AI tools can fully support the wide range of English repertoires in our world, much less supporting multilingual repertoires. On the other hand, GenAI erases variation and flattens nuances, especially when reviewers, editors, or other stakeholders in the AWP space treat divergence from dominant norms as error rather than evidence of legitimate linguistic diversity, which would continue to entrench epistemic injustice where alternative linguistic norms are devalued under the guise of correctness (Fricker, 2007). To realise this potential of GenAI to promote linguistic inclusivity, it is recommended that AI systems should be co-designed with input from linguists, technocrats and community people, including those from Indigenous, translingual/multilingual, migrant and Global South backgrounds. Interestingly, some studies have highlighted the possibility and affordances of this sort of co-creation in AI designs (Ofosu-Asare, 2025; Pinhanez & Wornyo, 2025), although these affordances still warrant further empirical interrogation. But, overall, there remain "doubts regarding their ability to develop the kind of rhetorical flexibility that is necessary for adapting writing to ever-changing contexts and demands" (Kuteeva & Andersson, 2024, p. 561).

At no point does this dialogue reject GenAI outright. Rather, it calls for situated, critical, and inclusive practices at multiple levels. This position views linguistic diversity not as a stylistic variable but as an ethical and epistemological concern, and as a matter of whose knowledge practices are validated or marginalised by automation. At the institutional level, contributors urge journals and universities to develop clearer, equity-informed guidelines on GenAI use for writers, reviewers and editors. At the individual level, contributors highlight both the risks of overreliance on GenAI and the possibilities of creatively adapting it to one's own linguistic identity. Iker sees promise in how scholars can "edit or completely rewrite GenAI-generated outputs... in ways that reflect their linguistic identity." This potential, however, must be supported by critical literacy and publishing cultures that value difference rather than erase it.

In conclusion, this scholarly dialogue shows that GenAI is neither inherently inclusive nor exclusive. Its impact depends on who trains it, who uses it, and under what institutional and ideological conditions it is used. The analysis here suggests that achieving equity in GenAI use requires developing the critical awareness of how tools mediate linguistic power and voice (Bender et al., 2021). The five contributors make a compelling case for reframing GenAI not just as a tool, but as a site of struggle over linguistic legitimacy, scholarly ethics, and global knowledge equity. Read alongside recent findings from interviews with applied-linguistics journal editors who highlight AI policy ambiguity, workload strain, and the urgent need for field-specific guidance on GenAI use (Moorhouse et al., 2025), this dialogue offers a complementary perspective by examining how such institutional guidance can shape the (il)legitimacy of non-dominant Englishes in AI-mediated publishing. In short, the policy question is inseparable from the sociolinguistic one: governance that overlooks varietal legitimacy risks automating older hierarchies under new tools. As institutions, journals, and researchers respond to the proliferation of GenAI use in AWP, they must ask not only how these tools can be used efficiently, but also for whom, by whom, and at what cost. Finally, the approach adopted in this paper affirms the value of creating dialogic spaces to include diverse perspectives. Despite the discussants not sitting together in one room to respond to these issues, the connections between their viewpoints indicate the shared views of sociolinguists on the role of GenAI in AWP. Nevertheless, this discussion is not exhaustive. As mentioned in the introduction, it reflects, to some degree, the enduring institutional concentration of scholarship in the Global North, even though some of the contributors bring Global South perspectives to bear on their analyses. Future dialogues would benefit from the inclusion of scholars and editors based in the Global South, whose experiences remain underrepresented in current AWP debates.

It is hoped that this collective reflection contributes to the ongoing discourse on inclusivity, policy, AI literacy, authorship, and linguistic justice in an age in which AWP is increasingly being shaped by artificial intelligence.